\documentstyle{llncs}
\input epsf

\begin{document}

\title{Learning to Play Pong using Policy Gradient Learning}

\author{Somnuk Phon-Amnuaisuk}
\institute{Media Informatics Special Interest Group, \\
Centre for Innovative Engineering, Universiti Teknologi Brunei,\\
School of Computing \& Informatics, Universiti Teknologi Brunei.
\texttt{somnuk.phonamnuaisuk@utb.edu.bn}
}

\maketitle

\begin{abstract}
Activities in reinforcement learning (RL) revolve around learning the Markov decision process (MDP) model, in particular, the following parameters: state values, $V$; state-action values, $Q$; and policy, $\pi$. These parameters are commonly implemented as an array. Scaling up the problem means scaling up the size of the array and this will quickly lead to a computational bottleneck. To get around this, the RL problem is commonly formulated to learn a specific task using hand-crafted input features to curb the size of the array. In this report, we discuss an alternative \emph{end-to-end} Deep Reinforcement Learning (DRL) approach where the DRL attempts to learn general task representations which in our context refers to learning to play the Pong game from a sequence of screen snapshots without game-specific hand-crafted features. We apply artificial neural networks (ANN) to approximate a policy of the RL model. The policy network, via \emph{Policy Gradients} (PG) method, learns to play the Pong game from a sequence of frames without any extra semantics apart from the pixel information and the score. In contrast to the traditional tabular RL approach where the contents in the array have clear interpretations such as $V$ or $Q$, the interpretation of knowledge content from the weights of the policy network is more illusive. In this work, we experiment with various \emph{Deep ANN} architectures i.e., \emph{Feed forward ANN} (FFNN), \emph{Convolution ANN} (CNN) and \emph{Asynchronous Advantage Actor-Critic} (A3C). We also examine the activation of hidden nodes and the weights between the input and the hidden layers, before and after the DRL has successfully learnt to play the Pong game.  Insights into the internal learning mechanisms and future research directions are then discussed.
\end{abstract}
\subsubsection*{Keywords}
Deep reinforcement learning; Policy gradient learning; Asynchronous Advantage Actor-Critic model

\section{Introduction}
The field of reinforcement learning (RL) has been an active research domain since the 1990s \cite{sutton17}. RL has been successfully applied to various problem domains including Game AI \cite{tesauro94,silver16,span11}. 
RL learns to behave in an environment, even though an imprecise evaluation of actions is not available, by randomly sampling the state space and remembering fruitful associations between the observed states and actions. Traditionally, a tabular approach is a common choice of implementation in various RL learning techniques e.g., \emph{temporal different learning (TD)}, \emph{Q-learning}, and \emph{SARSA}. 

The tabular RL approach works well when the state space is not too large because the RL can thoroughly explore the state space. Therefore, it is crucial that the environment be carefully abstracted to finitely enumerate all possible states.
Unfortunately, many real life problems have a large state space and it is likely that the RL could only partially learn its policy $\pi$. Such cases lead to the instability of the learning algorithms and a fluctuation in the performance of the tabular RL approach due to poor experiential exposure during the learning stage. The issue could be mitigated if the balance between exploitation and exploration is maintained  and enough time is given during the learning stage. The balance between exploring uncharted states and exploiting fruitful states ensures that near optimal policy is learnt within a reasonable time. This is still an open research issue and recent progress in this issue has been the application of \emph{deep learning} (DL) to RL.

Interest in applying deep learning to reinforcement learning has surged in the last decade. This may be contributed to the recent advances in DL \cite{lecun15} and the open-source movements which promote accessibility to new concepts/algorithms for a wider group of researchers\footnote{See \texttt{http://karpathy.github.io/2016/05/31/rl/}}. Deep Reinforcement Learning (DRL) applies DL techniques to approximate one of the following RL components: state values, $V$; state-action values, $Q$; policy $\pi$ or the model of the Markov decision process (MDP). 
The success story of Deep Q-Network (DQN) \cite{mnih15} and the introduction of many game test-beds such as AI Gym and Arcade Learning Environment (ALE) \cite{bellemare13} have revived interests in AI-game research, especially the application of DRL to learn how to play a game without explicit representations of game-play strategies, aka the end-to-end approach \cite{kriz12,sukhbaatar15}. The end-to-end approach expects the system to learn important characteristics without explicit input from a human programmer.  

The domain here is the \emph{Atari Pong game} from the \emph{AI Gym} implementation (https://gym.openai.com). The Pong game is a simple two-player turn playing game, where a player character (PC), which is the RL agent, and the non-player character (NPC), which is the AI gym, can take one of the following three actions: no movement, move the paddle up or move the paddle down. We are interested in the end-to-end approach where pixel information of the Pong game are taken as input without manually converting them into other features. The input to our DRL system is just the  pixel information  and the score after each served ball. 
The system learns a policy $\pi(a|s; w)$ without any direct supervision, i.e., whatever action $a$ that is preferred given a state $s$. The learning is obtained by adjusting the neural network weight $w$.

We are intrigued by some salient properties of the DRL approach such as (i) the ability to handle a large state space, and (ii) the appearance of general learning framework without explicit hand-crafted features. This motivated us to explore the DRL with the following techniques: the feed-forward neural network (FFNN), the convolution neural network (CNN) \cite{sermanet12}, and the asynchronous advantage actor-critic agents (A3C) \cite{mnih16}.
We hope to gain some insights into what has been learnt by examining the activation patterns of the hidden layer nodes, patterns of weights in the policy network, etc., after the policy network has successfully learnt to play the Pong game.

The rest of the paper is organized into the following sections: Section 2 discusses the background of DRL; Section 3 presents a formal description of the Pong game in the DRL framework; Section 4 discusses the policy gradient learning using FFNN method; Section 5 discusses the policy gradient learning using A3C method; Section 6 discusses the results and analysis of the neural network weights, as well as the activations of the hidden layer; and finally, the conclusion and suggestions for further research are presented in Section 7.

\section{Background on Reinforcement Learning and Deep Learning}
Game AI researchers have been working on behaviour learning issues for 
decades \cite{span11,hwang10,hausk14,diwang15}. The introduction of Arcade Learning Environment (ALE) \cite{bellemare13} has revived interest in the application of reinforcement learning technique in behaviour learning issues.
Traditionally, three popular tactics are employed in tabular reinforcement learning implementations: TD, Q-learning and SARSA \cite{sutton17}. It is called tabular RL since the state values or state-action values are commonly implemented as tables (a 2D array). For example, in the Pong game, the Cartesian product of the positions of the two paddles, the positions of the ball and the control of the player's paddle express all possible states. When the size of the lookup table is small enough, all entries in the table can be reached through experiential updates and the tabular approach could successfully learn the optimal solution of the task.

In \cite{mnih15}, the convolutional neural network was first employed to represent the game states of the following Atari-2600 games: Pong, Breakout, Space Invaders, Seaquest and Beam Rider. The representation of game states using visual information (i.e., screenshot pixels) creates a large amount of states. In the Pong game, the number of possible states from a screenshot of size 80 $\times$ 80  pixels can go up to $2^{6400}$ states (assuming that each pixel represents only two possible states). A tabular approach will not be practical with the problem of this size. Here, we resort to a DRL approach using artificial neural network to encode policy $\pi$ and employ a policy optimization approach to approximate the $V, Q$ or $\pi$ parameters. The action policy $\pi$ is encoded as weights in a policy network $\pi(w)$. The policy network  does not represent an approximate value functions (e.g., $V, Q$) as a table but as a parameterized functional form using a computational model such as artificial neural networks \cite{williams92}.

There are two main umbrellas for RL policy network learning approaches: a gradient-free approach and a gradient-based approach. The gradient-based approach employs gradient information obtained from the search landscape to guide its search. This can be obtained by establishing a loss function or by perturbing the state vector which can be seen as sampling around the local current state. From the set of perturbed samples, the gradient can be computed.
The gradient-free approach optimizes the policy $\pi$ using examples randomly sampled from the state space. The gradient-free learning approach may be guided by mechanisms such as \emph{evolutionary strategy} \cite{span15} 
or other search heuristics. 
The gradient-free approach can work well when the number of optimized parameters are small since the state space is heuristically explored. For a problem with a large state space, performance may suffer from non-exhaustive experience. Hence the gradient-based approach is preferred \cite{silver14,andry16}. 

\subsection{Open Research Issues}
The application of DL to RL opens up many new research issues which are not fully understood yet. From our perspective, the following two important issues are important characteristics of DRL: (i) how to speed up the learning process? and (ii) what is learnt in a policy network?

Speeding up the learning process and striking a balance between exploration and exploitation are the common research themes in RL and DRL communities. The recent Asynchronous Advantage Actor-Critic tactic (A3C) \cite{mnih16} effectively utilizes many agents working on the same environment. The term \emph{advantage} refers to the use of a biased reward $R$ - $V(s)$ instead of a bare reward $R$ \cite{schulman15}. The term \emph{asynchronous} refers to the fact that different agents independently explore the environment, and the term \emph{actor-critic} refers to the mechanism in A3C that allows agents to exploit shared knowledge through the critic component. The sharing of knowledge yields a positive effect in speeding up the learning process and striking a balance between exploration and exploitation. 

It is interesting to ask what is learnt by the policy network when DRL is fully trained. In the tabular RL, state values or state-action values are captured in table entries. 
It is, however, less obvious how the association states and actions are captured in the weights of the policy network.
It has been observed that a successfully trained convolution neural network (CNN) in the end-to-end image recognition task could reveal a hierarchical structure of features (simple features at the near input layer such as ridgelet, circular, curvature, grating, etc., to complex features at the output layer such as eyes and other meaningful labels). There is a clear increment in semantic content \cite{zeiler13,yosinsky15,zhou17}.
This automated abstraction of conceptual information is, however, not well understood yet.

\section{Formal Description of Pong for Policy Gradient learning}
Policy gradient learning differs from the tabular state values learning or the tabular state-action values learning approach since tabular approach updates its table entry by entry. The tabular approach is inefficient if the state space is large. The large state space often leads to a poor generalization of ability since the whole state space cannot be thoroughly explored. In contrast, policy learning efficiently approximates the policy by back-propagating errors to adjust the weights ${W}$ of the policy network. It learns all policies at once and converges them to a sub-optimal policy.

In the Pong game, an episode is complete when one of the players obtains 21 points. Hence in one episode, a sequence of image frames ${s}_t \in {\mathcal{R}}^{high \times width}$, a sequence of actions ${a}_t \in \{1,2,3\}$ that the RL agent has taken\footnote{In the Pong game, these actions are the three paddle actions: up, down and still.} and a sequence of reward signals $r_t \in \{-1,0,1\}$ constitute a batch of experience for that episode. Since the RL is playing against an NPC, if the NPC wins then the evaluative reward, at that frame $t$, will be $r_t = -1$, and if RL wins then $r_t = 1$. However while the game is still running, the reward $r_t = 0$. Hence, there will be thousands of frames with a lot of evaluative reward $r_t$ = 0 and occasionally with $r_t$ -1 or 1. The evaluative information at time $t$ must be propagated to earlier actions, $t-1$ since the merit or the penalty obtained at time $t$ are also contributed by previous actions. This is obtained by propagating the reward along the historical steps of the current game: $r_{t-1} =  r_{t-1} + \gamma r_{t} $, where $\gamma$ is the discount factor. The accumulative $R$ is 
\begin{equation} R = \sum_{n=0}^N \gamma^n r_{n}  \end{equation}
\begin{figure}[!ht]
\begin{center}\leavevmode
\epsfxsize=11cm
\epsfbox{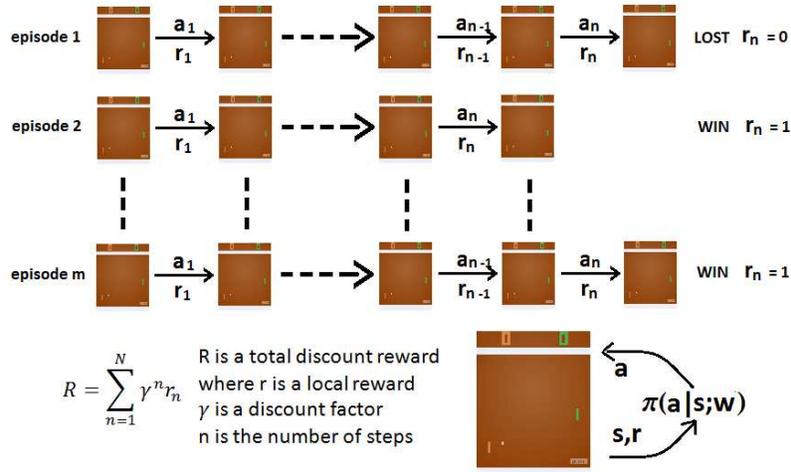}
\end{center}
\caption{Describing the Pong Game from the DRL perspective}
\label{pongRL}
\end{figure}

\section{Application of Policy Gradient learning in FFNN} \label{singleagent}
\begin{figure}[!ht]
\begin{center}\leavevmode
\epsfxsize=9cm
\epsfbox{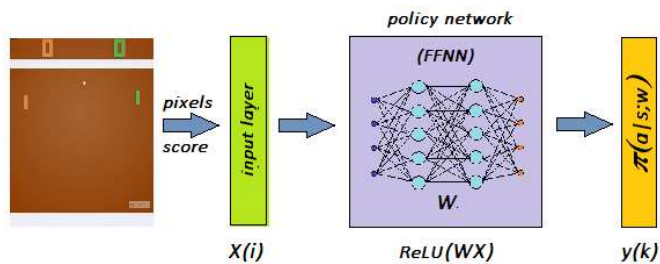}
\end{center}
\caption{An end-to-end RL implements a policy gradient network using feed-forward neural network. The network takes a pixel information as input and suggests a player's action output. The network may be constructed with one or more hidden layers.}
\label{pong}
\end{figure}

Let us formulate the Pong game from the end-to-end RL perspective \cite{levine16}; let ${s}$ denote an input vector constructed by flattening a 2D array of input pixels, let $\pi(a|s;w)$ denote a policy function which predicts action $a$, given a state $s$ and the parametrized weight $w$. The predicted ${y}$ from the policy network indicates the best action according to the policy $\pi$. The output $y$ is encoded in the \emph{one-hot} encoding fashion. Finally, let  ${W^l_{ij}}$ be a weight matrix of layer $l$ connecting the nodes from the input set $i$ to the output set $j$, as illustrated in Fig \ref{pong}.

After the RL has learnt to play the Pong game with a policy $\pi(w)$, we expect the accumulated return, ${R}$ to be maximized.
\begin{equation} R = \sum_{t=0}^H \pi(a_t|s_t; w) r_t  \end{equation}
The above equation maximizes $R$ if the probability of taking an action $a_t$ and the evaluative $r_t$ (after the reward propagation) have high values. In other words, it tells us that $R$ is maximized if the right decision is taken. The decision for these actions are conditioned according to the state and the parameterized neural network weights $w$. 

Without the loss of generality to the number of hidden layers, let us consider a neural network with just one hidden layer having the weights, $W^1_{ij}$ and $W^2_{jk}$, connecting between the input layer to the hidden layer and the hidden layer to the output layer (see Fig. \ref{pong}), let the activation functions of layer $W^1$ and $W^2$ be the ReLU($\cdot$) and Sigmoid($\cdot$) respectively, and let $x_i$ be a vector constructed by flattening the differences $s_t - s_{t-1}$. The differences between the two frames capture the temporal information. The predicted action $y_k$ can be expressed as:
\begin{equation} y_k = Sigmoid(\sum_j[\;ReLU(\sum_i [x_{i}\;W^1_{ij}])\; W^2_{jk}])  \end{equation} 
The discrepancies between the output $y_k$ and the \emph{biased random label} $y_{k}'$ form the loss function:
\begin{equation} loss = (y_k - y'_k)^2  \end{equation}
where $y_k$ and $y'_k$ are the one-hot representation of an action $a$ and a corresponding label respectively. It should be noted that the random labels are biased toward the actions recommended by the network. Hence, this can be seen as an \emph{on policy} learning with occasionally random actions.

The loss function above is optimized when the actions taken  agree with the labels. However, this does not imply optimal policy. The optimal policy is the policy that maximizes the accumulative return $R$ over trajectories $\tau$ where each trajectory is a sequence of state-action pairs i.e., $\tau = [(s_1,a_1), (s_2,a_2),...,(s_n,a_n)]$ in each episode. Hence the neural network weights $w$ condition the actions and rewards. We wish to find $w$ that maximizes the summation below
\begin{equation} \Sigma_{\tau} \pi(\tau; w) R(\tau) \end{equation}
The optimal weight is obtained by calculating derivative with respect to the $w$ to the above equation. This gives the likelihood ratio policy gradient as
\begin{equation} 
\begin{tabular}{rl}
$g \approx $ & $\nabla_w \ \Sigma_{\tau} \pi(\tau; w) R(\tau)$ \\
&\\
$ \approx $ & $\Sigma_{\tau} \  \frac{\pi(\tau; w)}{ \pi(\tau; w)} \nabla_w \pi(\tau; w) R(\tau)$ \\
&\\
$\approx $ & $\Sigma_{\tau} \pi(\tau; w)\nabla_w log \pi(\tau; w) R(\tau)$ \\
\end{tabular}
\end{equation}
The $\Delta w$ weight update below is derived from the above gradient:
\begin{equation} \Delta w =  \alpha [\nabla_w log \pi(a_t|{s}_t; w) (R-b)] \end{equation}
where $\alpha$ is the learning rate and $(R - b)$ is an advantage of the action $a_t$ over the expected base line\footnote{In this implementation, the base line is computed by averaging the discounted reward from each episode.} $b$. This is one of the variations of the REINFORCE algorithm \cite{williams92,tsitsiklis97}. In essence, the update increases the probability of an action having more advantage over other actions. Table \ref{pseudocode} gives a pseudo code of the policy gradient learning process.

\begin{table}
\begin{small}
\caption{Pseudo Code for Policy Gradient Learning}
\vspace{0.1cm}
\hrule 
\vspace{0.1cm}
Policy gradient method 
\vspace{0.1cm}
\hrule 
\vspace{0.1cm}
{\bf input}: frame pixels $s_0$ and the game score \\
\hspace*{0.8cm} a differentiable policy parameterization $\pi(a|s;w)$\\
{\bf output}: \emph{policy network weights} $w$\\
Initialize \emph{policy network weights} $w$\\
{\bf for} {\emph episode = 1 to M} {\bf do}\\
\hspace*{0.5cm}{\bf a,r} $\leftarrow$ Generate an episode $s_0,a_0,r_0,s_1,a_1,r_1,...,s_t,a_t,r_t$ following $\pi(\cdot|\cdot;w)$\\
\hspace*{0.5cm}{\bf a'} $\leftarrow$ Generate labels for the episode(labels are biased toward {\bf a}) \\
\hspace*{0.5cm}define loss from {\bf a} and {\bf a'}, loss = $(y_k - y'_k)^2$\\
\hspace*{0.5cm}for each t from t=T to t=0 of each episode:\\
\hspace*{1.0cm}$r_{t-1} =  r_{t-1} + \gamma r_{t} $\\
\hspace*{0.5cm}$w \leftarrow w + \alpha [\nabla_w log \pi(a_t|{s}_t; w) (R-b)] $\\
{\bf endfor}
\label{pseudocode}
\end{small}
\end{table}

\subsection{Results from Feedforward Neural Networks}
In the end-to-end setting, the input to the system are the pixels from the game, the score and the episode-complete flag. There are many tuning parameters. Since each experiment with FFNN takes days to complete (at least 40 hours for 20,000 episodes in our computer), variations of selected parameters are reported here. In this simulation, the numbers of hidden nodes, hidden layers and the learning rate are varied. Table \ref{param} summarizes the settings employed in our simulation. 
\begin{figure}[!ht]
\begin{center}\leavevmode
\epsfxsize=6.7cm
\epsfbox{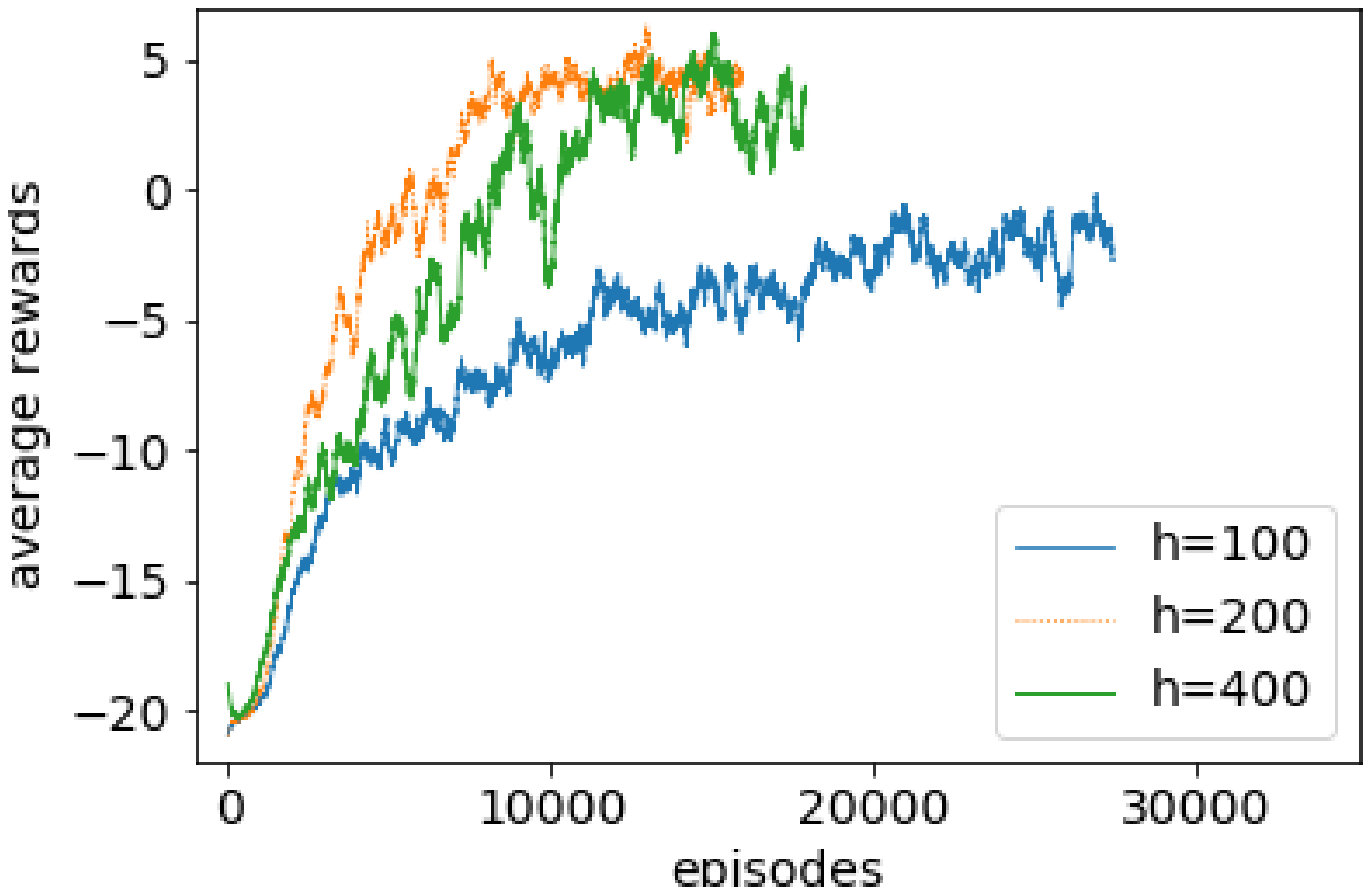}
\epsfxsize=6.7cm
\epsfbox{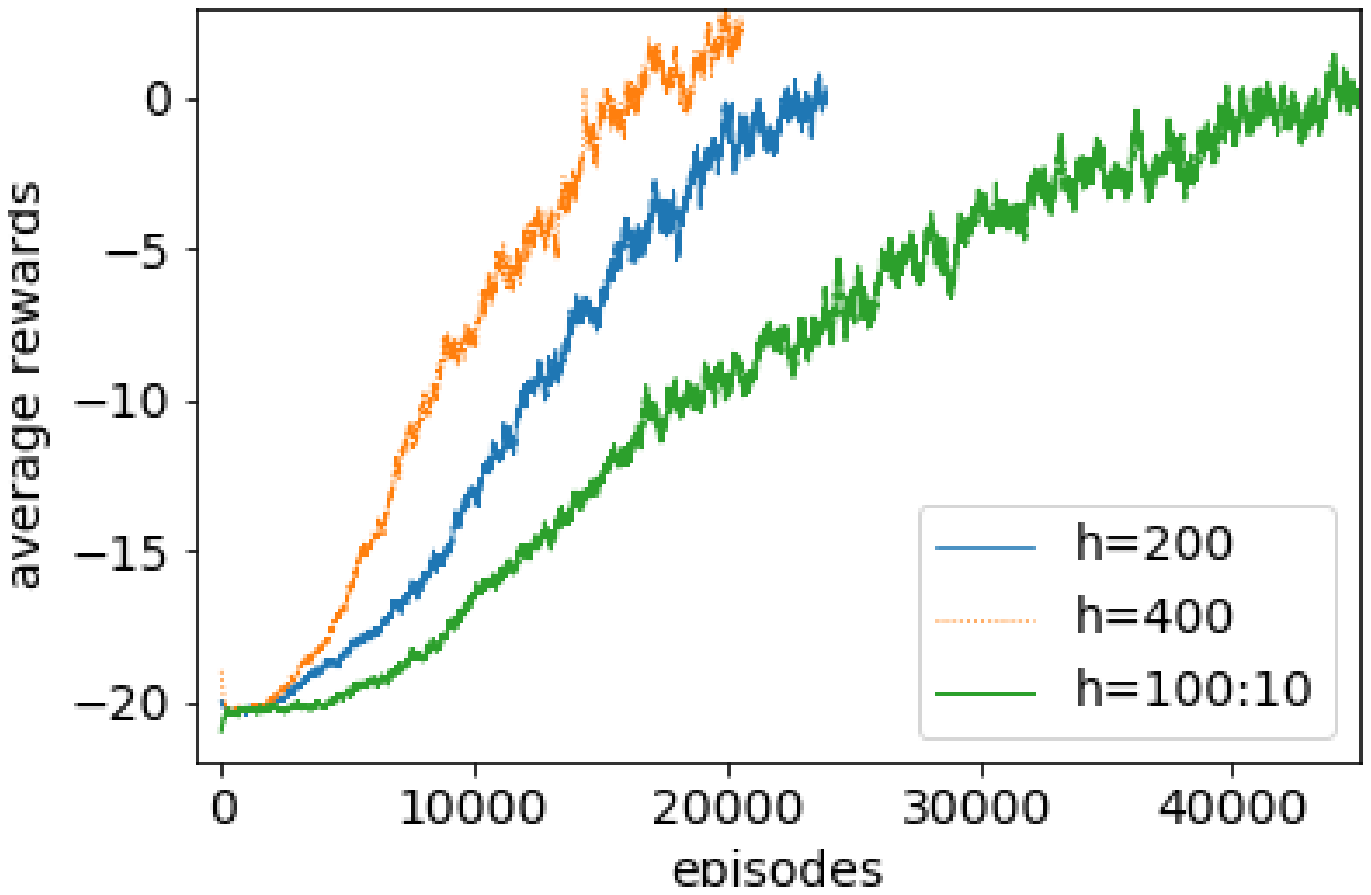}
\end{center}
\caption{Left: Summary of average scores observed from the networks with 100, 200, and 400 hidden nodes (single hidden layer), with the learning rate set at 0.001. Right: Summary of average scores observed from the networks with 100:10, 200, and 400 hidden nodes (100:10 denotes two hidden layers), with the learning rate set at 0.0001.}
\label{sim1-2}
\end{figure}

\begin{table}[htb]
\begin{scriptsize}
\caption{Control Parameters}
\begin{center}
\begin{tabular}{l|c|c|c|c} \hline \hline
{\bf Information} & \multicolumn{3}{c|}{\bf 1 hidden layer} & {\bf 2 hidden layers}\\ \hline
Input frame size & 80 $\times$ 80 & 80 $\times$ 80 & 80 $\times$ 80 & 80 $\times$ 80 \\
Hidden nodes & 100 & 200 & 400 & 100:10\\
Output nodes & 3 & 3 & 3 & 3\\
Learning rate $\alpha$  & 0.001 & 0.001,& 0.001,& \\
  & & 0.0001 & 0.0001 & 0.0001 \\
Discount factor $\gamma$  & 0.9 & 0.9 & 0.9 & 0.9 \\
Policy network & 6400:100:3 & 6400:200:3 & 6400:400:3 & 6400:100:10:3\\
Activation functions & ReLU & ReLU & ReLU & ReLU \\ 
 & Sigmoid & Sigmoid & Sigmoid & ReLU \\  
 & & & & Sigmoid \\  \hline
\end{tabular}
\label{param}
\end{center}
\end{scriptsize}
\end{table}

Figure \ref{sim1-2} shows the average scores of different network architectures and different learning rates. It is conclusive that learning does take place in both single-hidden layer, or two hidden-layer cases since the average scores of all runs increase with episodes. However, the behaviors of the learning algorithm are dependent on many factors such as weight initialization, learning rate, lost function, etc. Although it is clear that the policy gradient learning approach can successfully train the policy network, it is quite hard to draw any conclusive argument regarding optimal parameter-setting from the data that we have. 

\section{Application of Policy Gradient learning in A3C}
It has been shown in the previous section that policy gradient technique can successfully train the FFNN to reach a competitive level to or better than the computer. However, the training process takes quite a long time (after approx 20,000 episodes). It is desired to speed up this training process. In a recent work by \cite{mnih16}, the authors propose the \emph{Asynchronous Advantage Actor-Critic} (A3C) framework. It has been shown that knowledge from multi-agents can be combined together and this tactic has sped up the learning time and improved learning performance considerably. This concept reminisces the multi-agent framework \cite{panaluke05,span09} and the island model in the evolutionary computing \cite{zbigniew05} where knowledge sharing is one of the important mechanisms.
\begin{figure}[!ht]
\begin{center}\leavevmode
\epsfxsize=11cm
\epsfbox{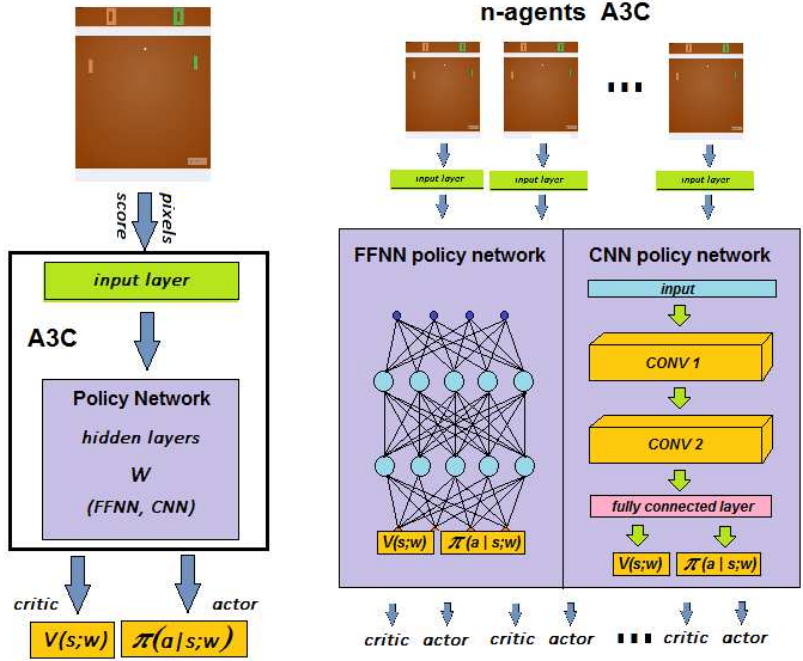}
\end{center}
\caption{Left pane: An end-to-end A3C implements a policy gradient network using feed-forward neural network (FFNN) and convolution neural network (CNN). The network takes the pixel information as input and suggests a player's action output as well as the quality of the state-action. Right pane: many agents play different games using the same policy network, experiences from different agents are employed to adjust the same policy network.}
\label{A3Carchitecture}
\end{figure}
\begin{table}[htb]
\begin{scriptsize}
\caption{Control Parameters A3C}
\begin{center}
\begin{tabular}{l|c|c|c} \hline \hline
{\bf Information} & {\bf FFNN 1} & {\bf FFNN 2 }  & {\bf CNN} \\ \hline
Input frame size & 80 $\times$ 80 & 80 $\times$ 80 & 80 $\times$ 80 \\
Output nodes & 3 & 3 & 3 \\
Learning rate $\alpha$  & 0.001 & 0.001 & 0.001 \\
Discount factor $\gamma$  & 0.9 & 0.9 & 0.9 \\ 
Activation functions & ReLU & ReLU & ReLU \\ 
Policy network & 6400:256:3 & 6400:1600:256:3 & 6400:Conv1:Conv2:256:3 \\ 
& & & Conv1 has 16 of 8$\times$8 kernel \\
& & & Conv2 has 32 of 4$\times$4 kernel \\ \hline
\end{tabular}
\label{param}
\end{center}
\end{scriptsize}
\end{table}

The main concept behind A3C is graphically explained in Figure \ref{A3Carchitecture}. 
The policy network of A3C has two output: (i) the actor provides an action $a$ and (ii) the critic provides the value, $V$ of the current input state. Three kinds of loss functions are derived from the action and the value: value loss ($\mathcal{L}_V$), entropy loss ($\mathcal{L}_H$), and policy loss ($\mathcal{L}_{\pi}$):
\begin{equation}  
\begin{tabular}{rl}
$\mathcal{L}_{\pi} =$ &  $- log\ \pi(a_t|{s}_t; w) (R-V(s;w))$ \\
&\\
$\mathcal{L}_V =$ & $ [R-V(s;w)]^2 $ \\
& \\
$\mathcal{L}_H =$ & $  - \Sigma [\pi(a_t|s_t;w) log\ \pi(a_t|s_t;w)]$ \\
&\\
Total loss = & ${\beta_1} {\mathcal L}_{\pi} + \beta_2 {\mathcal L}_V + \beta_3 {\mathcal L}_H$\\
\end{tabular} 
\end{equation}
where $\beta_n$ is arbitrary weight values\footnote{For example, ($\beta_1,\beta_2,\beta_3$) = (1.0, 0.5, 0.001). The entropy loss behaves as a regularization term and is commonly set at a small value.}. In this experiment, both FFNN and CNN \cite{sermanet12} are employed as the policy network. 

\begin{table}
\begin{small}
\caption{Pseudo Code for each actor-learner thread of A3C}
\vspace{0.1cm}
\hrule 
\vspace{0.1cm}
A3C Policy gradient method 
\vspace{0.1cm}
\hrule 
\vspace{0.1cm}
{\bf for} {\emph episode = 1 to M} {\bf do}\\
\hspace*{0.5cm}{\bf for} {\emph agent = 1 to n} {\bf do}\\
\hspace*{1.0cm}{\bf input}: frame pixels $s_0$ and the game score \\
\hspace*{1.0cm}{\bf output}: \emph{policy network weights} $w$\\
\hspace*{1.0cm}{\bf if} \emph{agent n still plays}\\
\hspace*{1.5cm}{\bf a,r,V} $\leftarrow$ play $s_0,a_0,r_0,s_1,a_1,r_1,...,s_t,a_t,r_t$ following $\pi(\cdot|\cdot;w)$\\
\hspace*{1.5cm}for each t from t=T to t=0 of each episode:\\
\hspace*{2.0cm}$r_{t-1} =  r_{t-1} + \gamma r_{t} $\\
\hspace*{1.5cm}Compute losses: value loss (${\mathcal L}_V$), entropy loss (${\mathcal L}_H$), and policy loss ($\mathcal{L}_{\pi}$)\\
\hspace*{1.5cm}${\mathcal L}(w) = \beta_1 {\mathcal L}_{\pi} + \beta_2 {\mathcal L}_V + \beta_3 {\mathcal L}_H$\\
\hspace*{1.5cm}$w \leftarrow w + \alpha [\nabla_w {\mathcal L}(w)] $\\
\hspace*{0.5cm}{\bf endfor}\\
{\bf endfor}
\label{pseudocodea3c}
\end{small}
\end{table}

\begin{figure}[!ht]
\begin{center}\leavevmode
\epsfxsize=9.5cm
\epsfbox{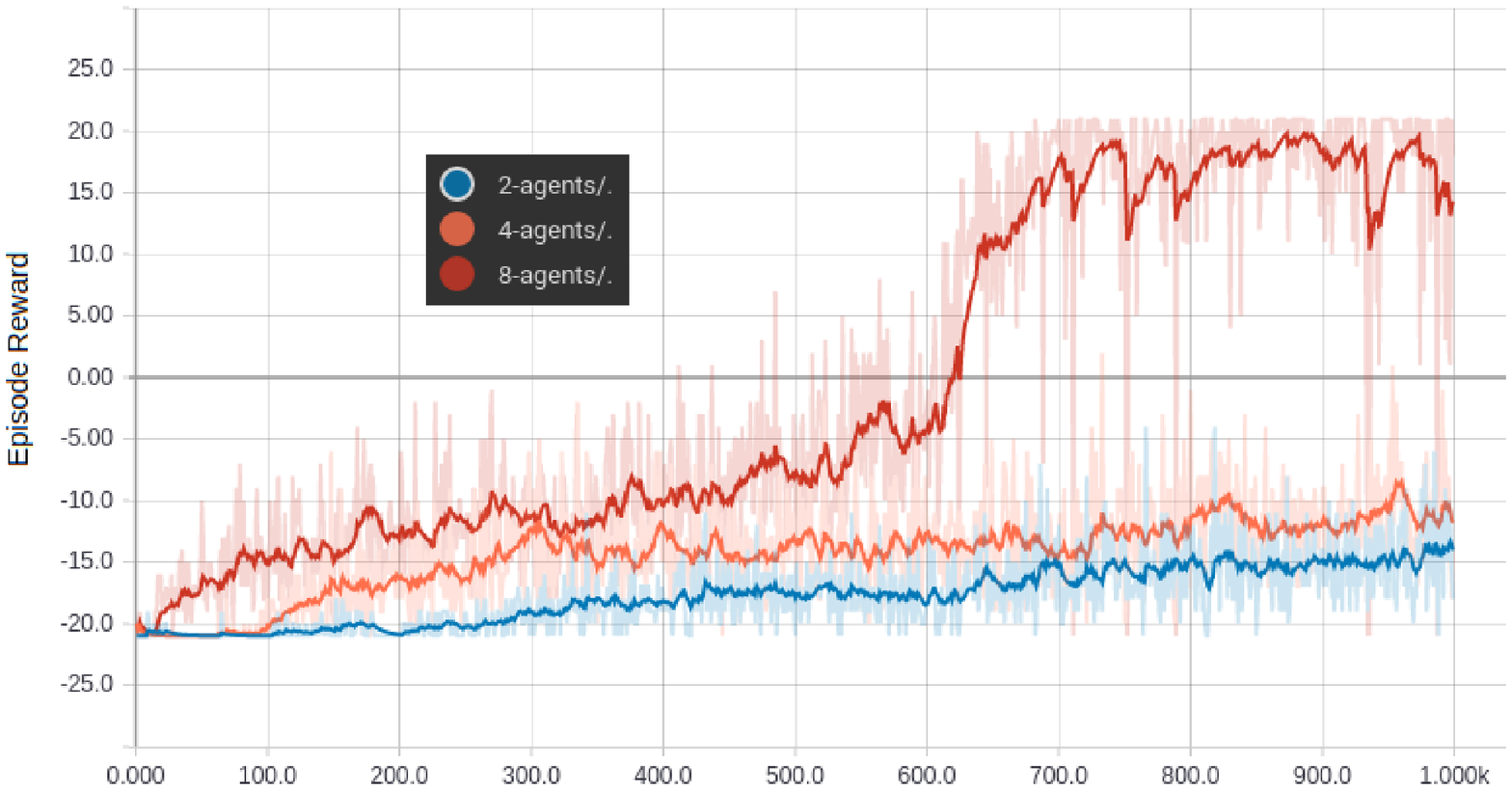}
\epsfxsize=8cm
\epsfbox{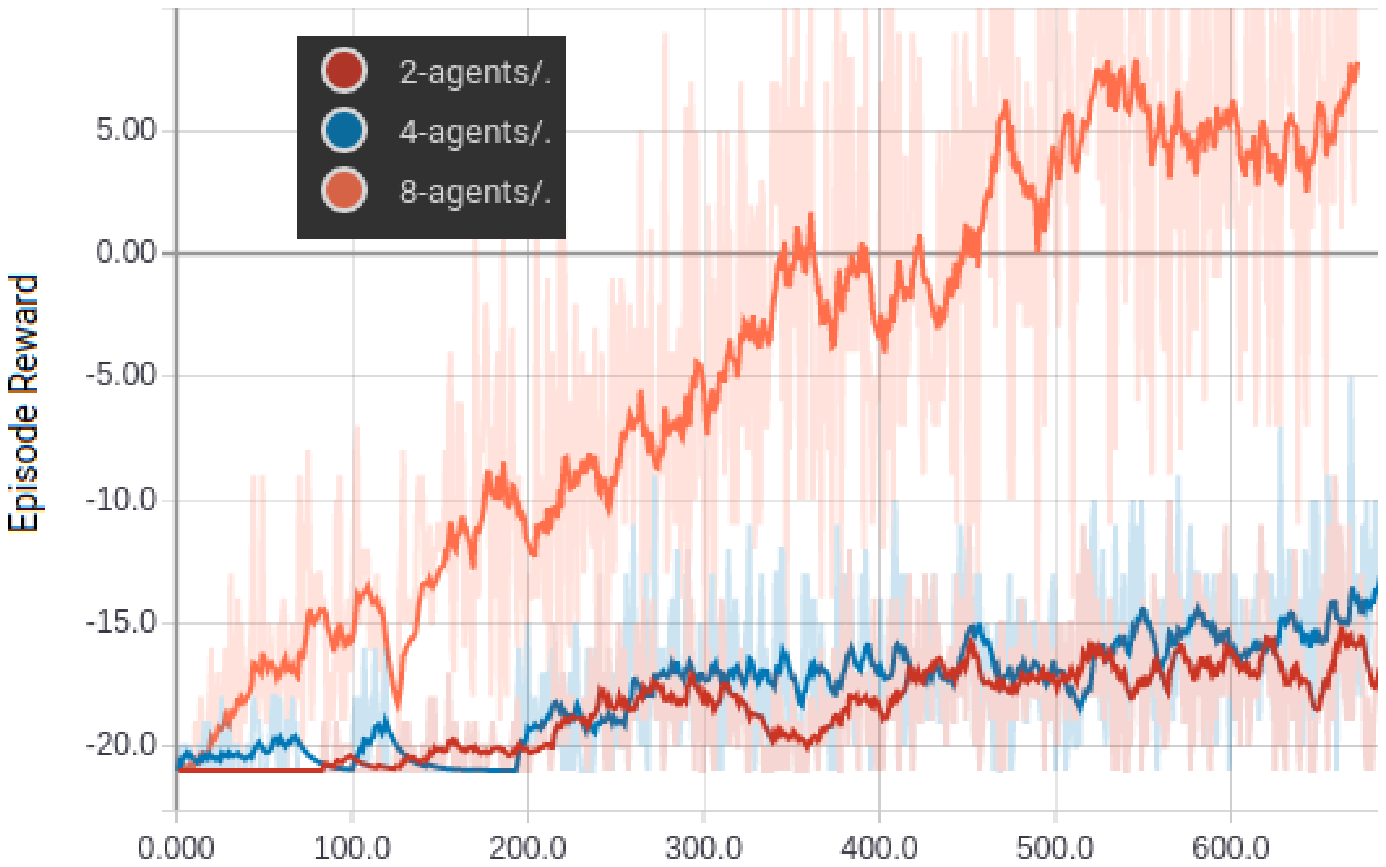}
\end{center}
\caption{A3C implements a policy gradient network using a single hidden layer FFNN (top pane) and  two hidden layers FFNN (bottom pane). The performance curves of the network running on 2, 4 and 8 agents are displayed. The y-axis shows the score, the x-axis shows the episode number.}
\label{FFNNa3c}
\end{figure}

\begin{figure}[!ht]
\begin{center}\leavevmode
\epsfxsize=9cm
\epsfbox{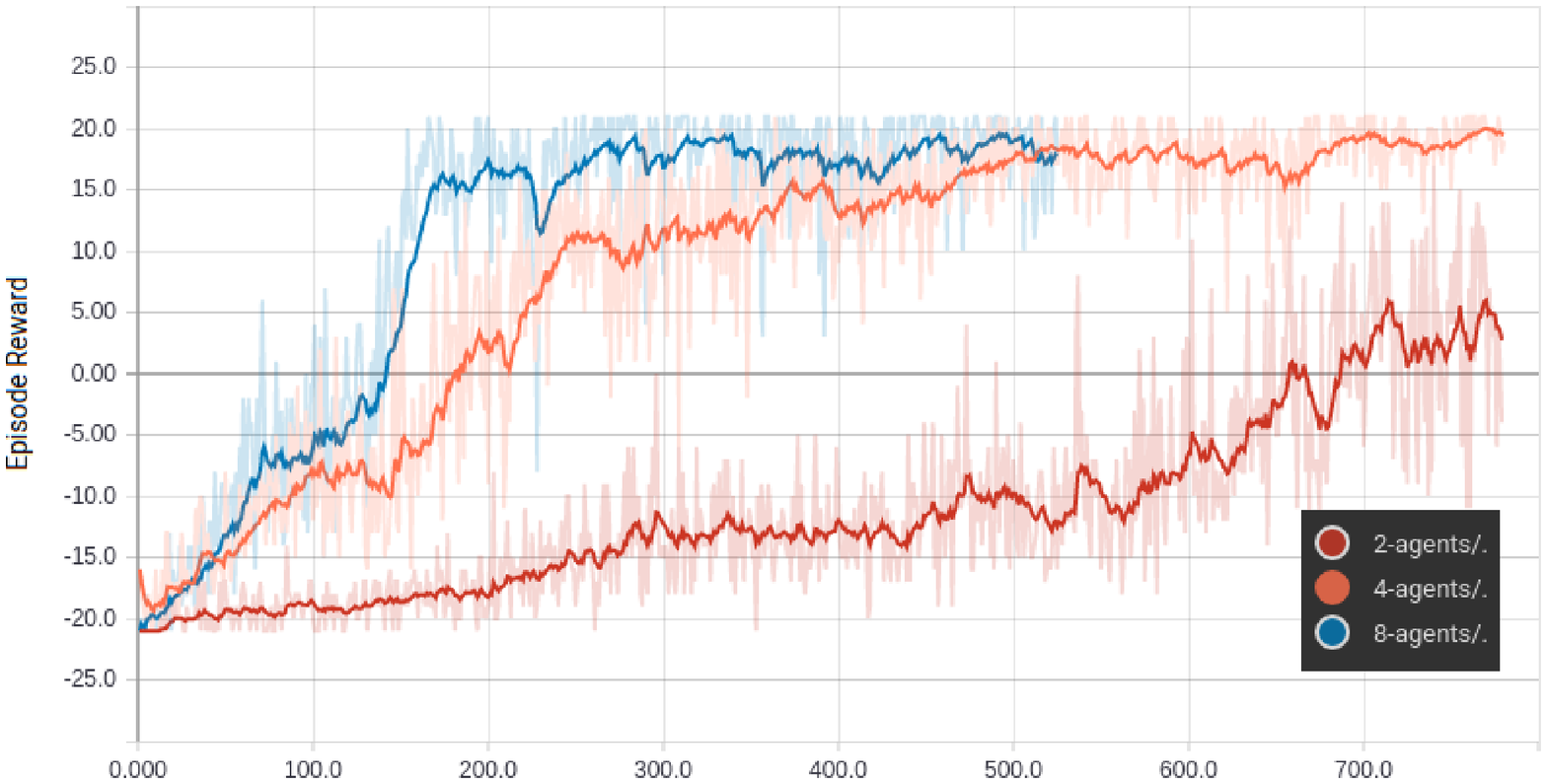}
\epsfxsize=9cm
\epsfbox{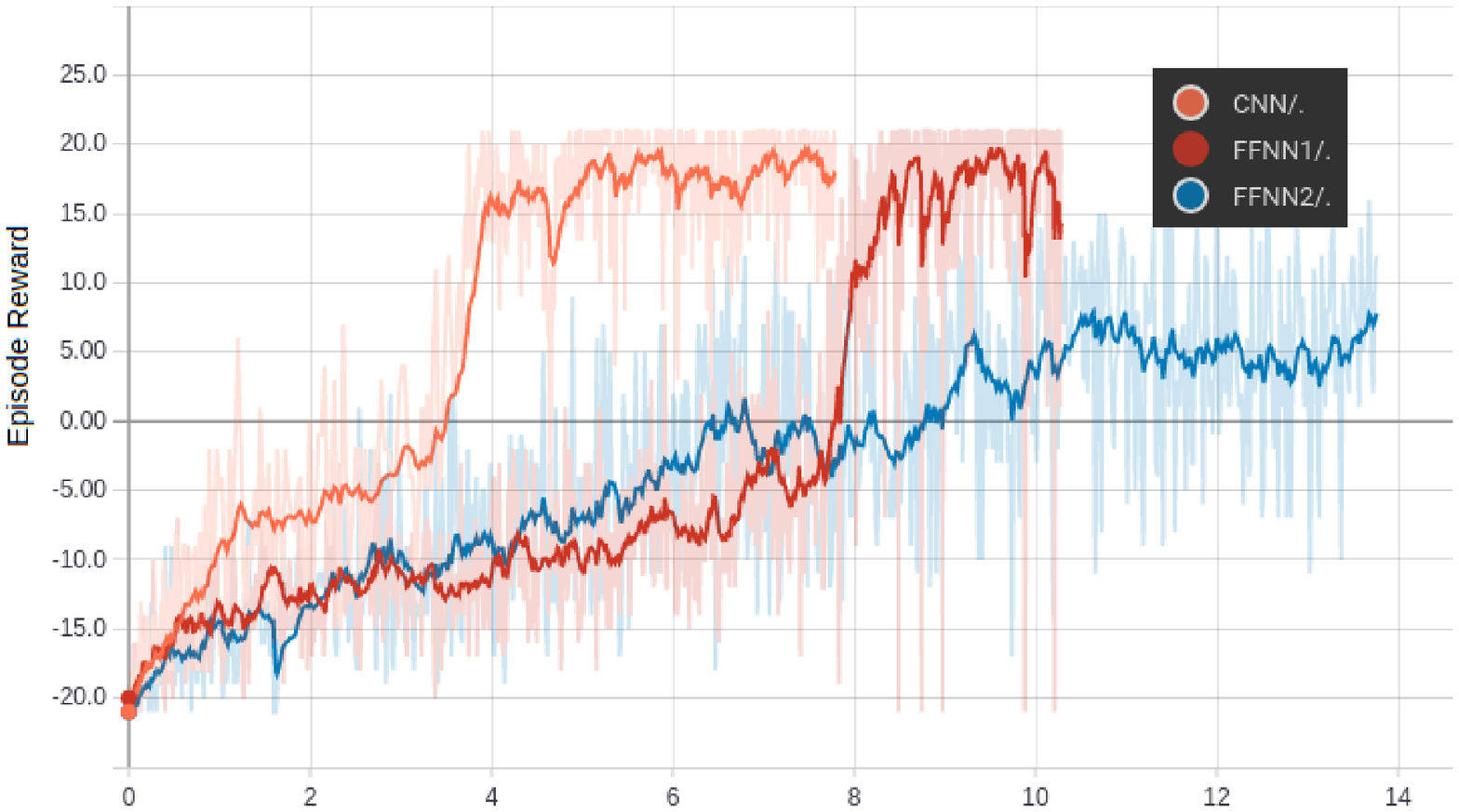}
\end{center}
\caption{Top pane: A3C implements a policy gradient network using convolution neural network. The performance curves of the network running on 2, 4 and 8 agents are displayed. The y-axis shows the score, the x-axis shows the episode number. Bottom pane: Performance comparison of three A3C networks implementing CNN, single hidden layer FFNN and two hidden layers FFNN. The CNN shows the best learning speed, followed by the single hidden layer FFNN and two hidden layers FFNN respectively. The y-axis shows the score, the x-axis shows the elapsed time (hrs).}
\label{CNNa3c}
\end{figure}

\subsection{Results from Asynchronous Advantage Actor-Critic (A3C)}
In A3C approach, each agent plays an independent game based on the same policy network. After each game, the policy network is adjusted based on the experiences of all the agents. Since all the agents solve the same problem, this can be seen as the collaborative exploration of all the agents on the different areas of the state space. Hence, this increases the speed and performance of the A3C method. 
The concept of A3C method is illustrated in Figure \ref{A3Carchitecture} (bottom pane). Table \ref{pseudocodea3c} gives a pseudo code of the policy gradient learning process.

Many games are simulated using different network architectures and different parameter settings.
The performance of the A3C method is more superior than a single agent approach carried out in section \ref{singleagent}. Figures \ref{FFNNa3c} and \ref{CNNa3c} show the performance of 2, 4 and 8 agents in each network. The learning rate of the CNN network clearly outperforms FFNN networks. The FFNN with a higher number of hidden layers appears to take a longer time to learn. Across the board, the higher the number of agents, the faster the learning rate.
\begin{table}[htb]
\begin{scriptsize}
\caption{Smoothed Score Performance Summary}
\begin{center}
\begin{tabular}{|l|c|c|c|c|c|c|} \hline \hline
\multicolumn{7}{|l|}{FFNN} \\ \hline
steps & 4000 & 8000  & 12000 & 16000 & 20000 & 24000  \\ \hline
h=100   &-9 &-7 &-5 &-4 &-3 & \bf{\emph{-2}}\\
h=100:10&-20 &-17 &-13 &-10 &-7 &\emph{-6}\\ \hline
\multicolumn{7}{|l|}{A3C-FFNN 1 hidden layer} \\ \hline
steps & 100 & 200  & 300 & 400 & 500 & 600  \\ \hline
2 agents&-21&-21&-19&-18&-17&\emph{-17}\\
4 agents&-20&-17&-13&-13&-13&\emph{-13}\\
8 agents&-15&-13&-11&-10&-8&\bf{\emph{-4}}\\ \hline
\multicolumn{7}{|l|}{A3C-FFNN 2 hidden layers} \\ \hline
steps & 100 & 200  & 300 & 400 & 500 & 600  \\ \hline
2 agents&-21&-21&-18&-18&-17&\emph{-16}\\
4 agents&-21&-20&-16&-16&-16&\emph{-15}\\
8 agents&-15&-11&-6&-3&3&\bf{\emph{5}}\\ \hline
\multicolumn{7}{|l|}{A3C-CNN}\\ \hline
steps & 100 & 200  & 300 & 400 & 500 & 600  \\ \hline
2 agents&-19&-18&-13&-12&-10&\emph{-8}\\
4 agents&-10&1&11&13&17&\emph{18}\\
8 agents&-7&15&16&17&19&\bf{\emph{20}}\\ \hline
\end{tabular}
\label{summary}
\end{center}
\end{scriptsize}
\end{table}

\section{Discussion}
\begin{figure}[!ht]
\begin{center}\leavevmode
\epsfxsize=10cm
\epsfbox{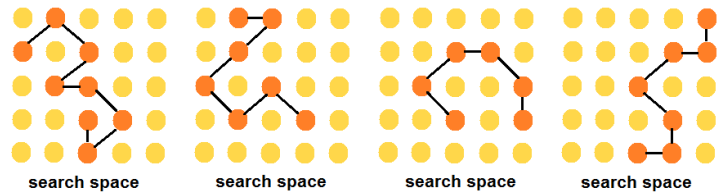}
\end{center}
\caption{Agents collaboratively search the search space and yield a faster convergence speed than the standard single agent approach.}
\label{a3cspeed}
\end{figure}

One distinctive feature of the A3C approach is its fast learning speed. We offer the graphics in Figure \ref{a3cspeed} to explain this. 
Let us view the search space as a set of states where each node represents a game state, where each agent starts the game at a different starting node and traverses the search space using its current policy $\pi$. In a single agent case, the policy $\pi$ is refined in each episode using the experience of a single agent. In the A3C case, since many agents collaboratively explore different parts of the same search space, unfruitful parts of the state space are pruned out early. The effect of pruning out the search space non-linearly speeds up the search and this results in a faster convergence rate of the learning process to the sub-optimal solution.

\subsection{Inspection of the Activations from the Hidden Nodes }
In order to examine the characteristics of the policy network, after it has been  successfully trained and able to beat the AI Gym, the weights of a trained network are frozen. Then the system is simulated for 50,000 steps and the activations of the nodes in the hidden layers and the output actions are recorded. These generate 50,000 pairs of hidden activation pattern and action pairs. An analysis of these pairs reveals the activation patterns for the up, down and still actions. 
\begin{figure}[!ht]
\begin{center}\leavevmode
\epsfxsize=9.5cm
\epsfbox{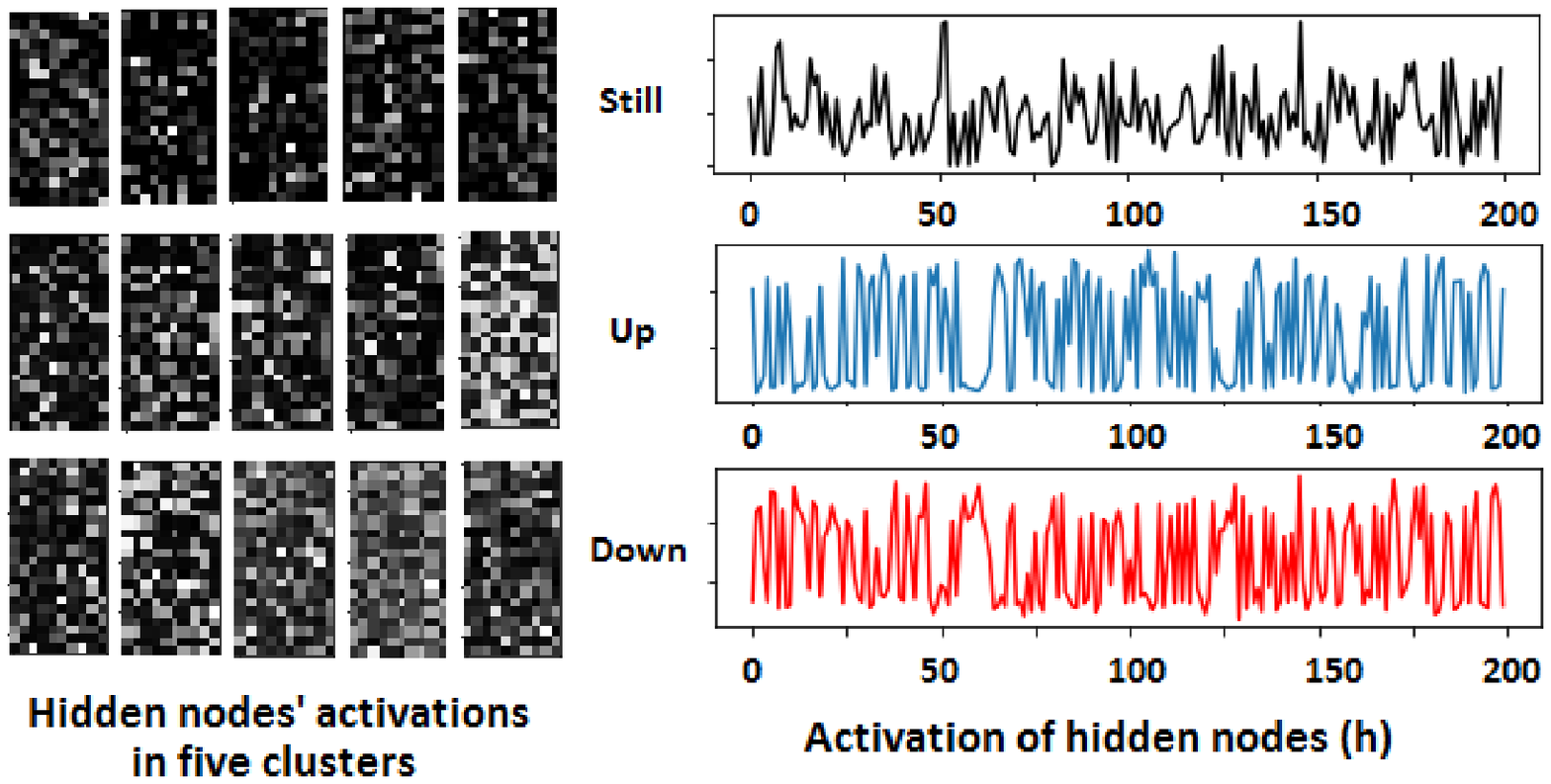}
\epsfxsize=12cm
\epsfbox{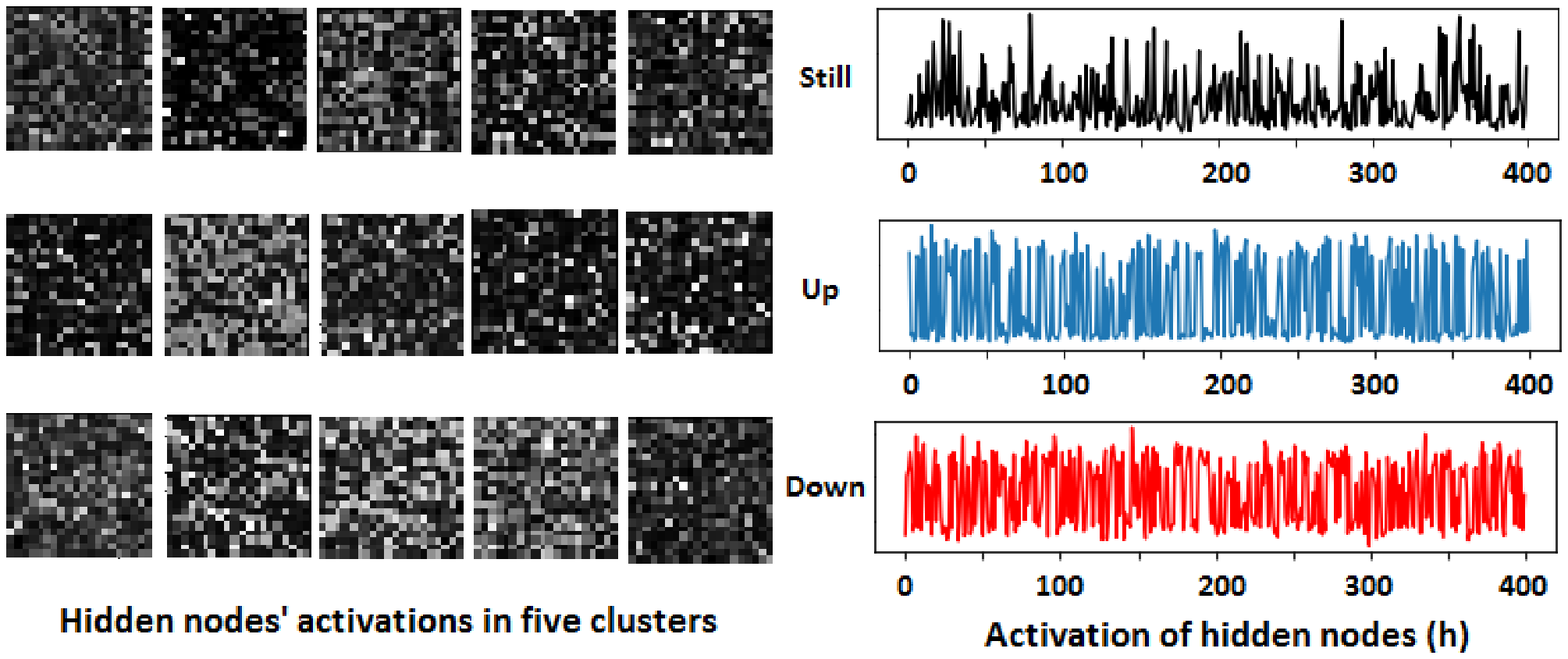}
\end{center}
\caption{Activation Patterns from 200 hidden nodes (top) and 400 hidden nodes (bottom). A clear grouping of nodes according to their functions is observed.}
\label{acthidden}
\end{figure}
Fig. \ref{acthidden} shows plots of hidden layer activations (top-right). The 200 hidden nodes are reshaped into a $20 \times 10$ image (top-left). It is observed that after the training, all the hidden nodes converge into three sets: nodes corresponding to the \emph {still} action, nodes corresponding to the \emph{up} action and nodes corresponding to the \emph{down} action. 

Among the \emph{still/up/down} nodes, it would be interesting to see further groupings inside each category. Here, the activation patterns in each \emph{still/up/down} group are further clustered into five clusters using k-means clustering.   Their results are shown in three rows; the first row shows five plots of activation strength corresponding to five clusters in the \emph{still} action, the second row for the \emph{up} action and the third row for the \emph{down} action.
The bottom pane of Figure \ref{acthidden} presents similar information but is from a network with 400 nodes in the hidden layer (it is reshaped into a $20 \times 20$ image).
The clear grouping of functionalities of these hidden nodes is biologically sound and has been reported in many places \cite{fentress76, getting80}

\subsection{Inspection of the Weight Patterns of the First Hidden Layer of FFNN}
\begin{figure}[!ht]
\begin{center}\leavevmode
\epsfxsize=10cm
\epsfbox{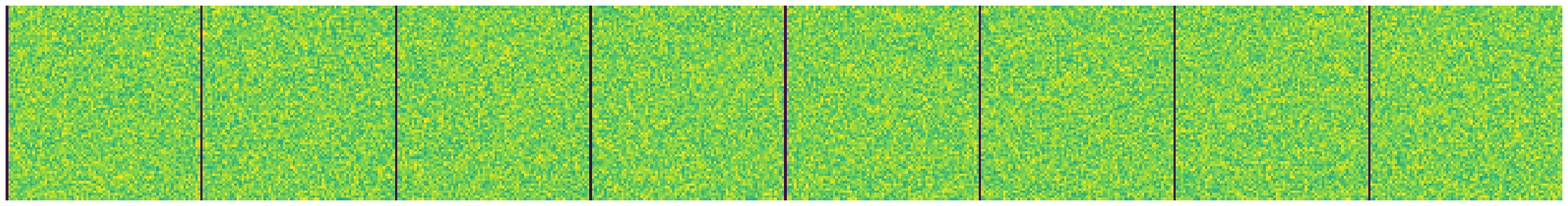}
\epsfxsize=10cm
\epsfbox{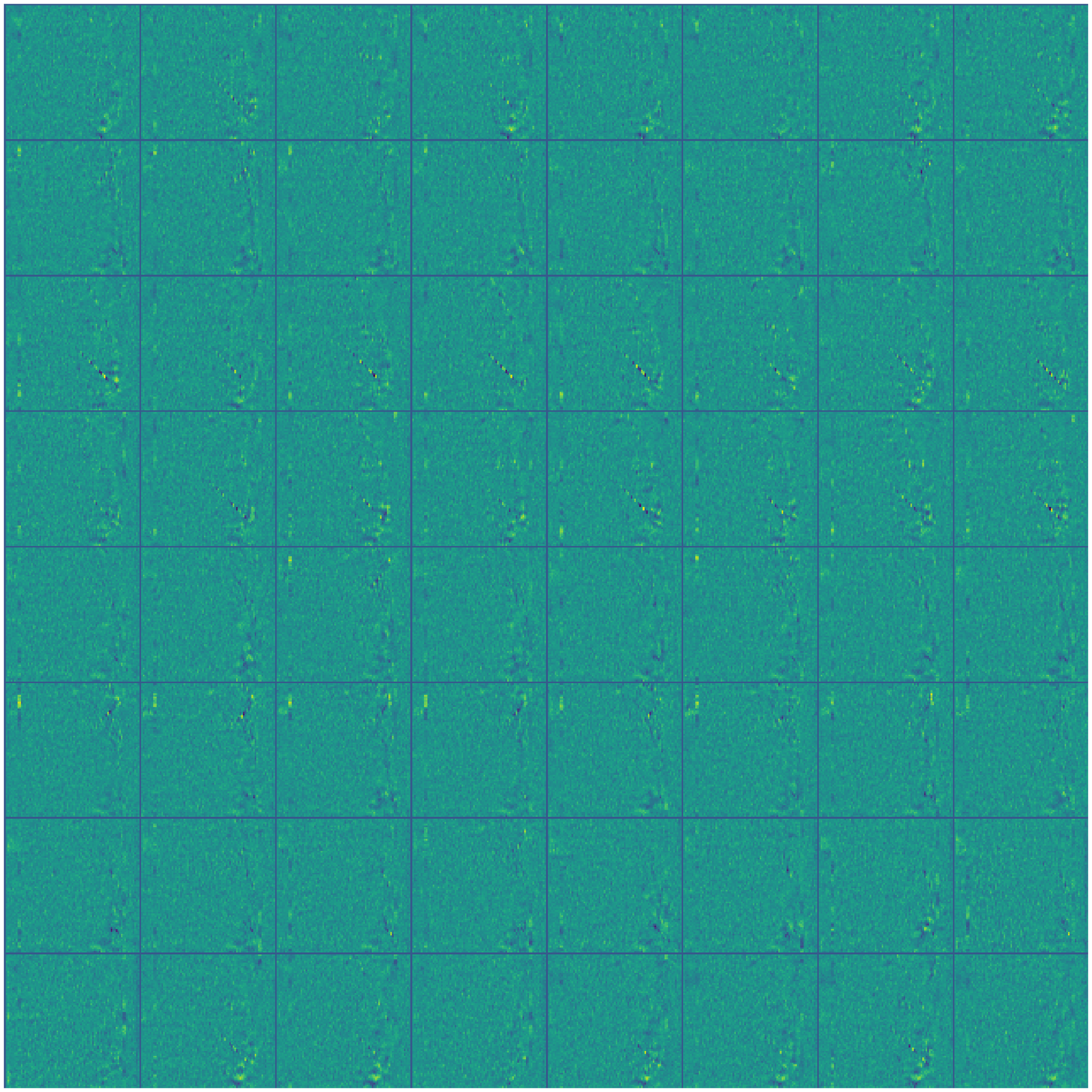}
\end{center}
\caption{This figure shows 72 plots (9 rows, 8 columns) of weights from some random hidden nodes of FFNN. The weights from input nodes $i$ to the hidden node $j$ are projected to form an imagery representation of a game board; (first row) before the learning, and (the bottom eight rows) after the learning. The trajectory of the paddlers and the ball can be clearly observed after the learning.}
\label{weight}
\end{figure}

Figure \ref{weight} shows the plot of weights $W_{ij}$ connecting input $x_i$ to the hidden nodes $h_j$. There are 6,400 input weights associated to each node in the first hidden layer. Representing these 6,400 weights as an $80 \times 80$ image reveals the game board which presents one of the effective visualization tactics \cite{zeiler13}. The first row of Fig. \ref{weight} shows the weights before the learning. No obvious pattern is observed. After the learning, a clear pattern, which could be seen as the trajectory of the ball, emerges such as shown in rows two and three. 
This means the node $h_j$ will produce high activation output, i.e., ReLU($\sum_i x_i W_{ij}$), if the corresponding pixel $x_i$ is active. The brighter the pixel, the higher the contribution to the hidden node activations. In the case of a single hidden layer, the action $y_k$ are computed from the contributions of the hidden nodes $Sigmoid(\sum_j h_j W_{jk})$. The weights $W_{jk}$ linearly combine the contributions from the different hidden nodes. In other words, the $W_{ij}$ learns the representation of the dictionary (can be seen as basis vectors). If the dictionary is complete, every input frame should successfully activate the appropriate hidden nodes, their appropriate combinations and therefore the appropriate actions that optimize the reward.

\section{Conclusion \& Future Direction}
The ANN has been applied as a function approximator for RL using hand-crafted features \cite{william88,bertsekas96}. This work investigates the learning function approximation in the end-to-end approach using deep reinforcement learning and without hand-crafted features. We show the results of policy gradient learning implemented using standard DRL and A3C methods. Their characteristics under different parameters are discussed. The following observations are conclusive from the results: (i) DRL is successfully applied to learn policy gradient; (ii) A3C approach shows a good learning speed since the policy network is explored in parallel; (iii) the activation of hidden layers has been successfully grouped according to functionalities: still, up, down; (iv) the weights between the input layer and the hidden layer show a strong correlation between ball trajectories and appropriate actions.

The above observations share parallel similarities to the dedicated functionality in different brain areas and the concept of Hebbian learning (i.e., observations iii \& iv). The policy gradient learning in DRL shares many similarities with learning behaviors observed in biological systems, for examples, the ability to learn from inexact feedback, and be able to adapt to changes in the environment. However, there are characteristics that seem to be difficult to realize with the current architecture. For example, does the policy $\pi(w)$ capture a deep gameplay strategy? Is it plausible for complex strategies to be captured by this style of learning mechanisms? Can a seemingly intelligent action performed by $p(a|s;w)$ be claimed as \emph{intuition}? We hope to investigate these sort of questions in our future work.

\begin{small}
\subsubsection*{Acknowledgments}
We wish to thank anonymous reviewers for their comments that have helped improve this paper. 
We would like to thank the GSR office  for the financial support given to this research. 
\end{small}

%
%


\begin{thebibliography}{}
\begin{small}
\bibitem{sutton17}
Sutton, R.S., Barto, A.G.: 
Reinforcement Learning: An Introduction (2nd Edition, online draft). MIT Press. (2017)

\bibitem{tesauro94}
Tesauro, G.: 
TD-Gammon, a self-teaching backgammon program, achieves master-level play. 
Neural Computation, 6(2):215-219 (1994)

\bibitem{silver16}
Silver, D., Huang, A., Maddison, C.J., Guez, A., Sifre, L., van den Driessche, G., Schrittweiser, J., Antonofglou, I., Panneershelvam, V., Lanctot, M., et al.:
Mastering the game of Go with deep neural networks and tree search.
Nature, 529(7587): 484-489 (2016)

\bibitem{span11}
Phon-Amnuaisuk, S.:
Learning chasing behaviours of non-player characters in games using SARSA. 
In: Proceedings of the International Conference EvoApplications. LNCS 6624  pp. 133-142 (2011)

\bibitem{lecun15}
LeCun, Y., Bengio, Y., Hinton, G.: 
Deep learning. Nature, 521:436-444 (2015)

\bibitem{mnih15}
Mnih, V., Kavukcuoglu, K., Silver, D., Rusu, A. A., Veness, J., Bellemare, M. G., Graves, A., Riedmiller, M., Fidjeland, A. K., Ostrovski, G., Petersen, S., Beattie, C., Sadik, A., Antonoglou, I., King, H., Kumaran, D., Wierstra, D., Legg, S., Hassabis, D.:
Human-level control through deep reinforcement learning. 
Nature, 518:529-533 (2015)

\bibitem{bellemare13}
Bellemare, M.G., Naddaf, Y., Veness, J., Bowling, M.:
The arcade learning environment: An evaluation platform for general agents.
Journal of Artificial Intelligence Research 47: 253-279 (2013)


\bibitem{kriz12}
Krizhevsky, A., Sutskever, I., Hinton, G. E. (2012). 
Imagenet classification with deep convolutional neural networks. 
Advances in Neural Information Processing Systems. pp. 1097-1105 (2012)

\bibitem{sukhbaatar15}
Sukhbaatar, S., Szlam, A., Weston, J., Fergus, R.:
End-to-end memory networks. 
In: Advances in Neural Information Processing Systems. pp. 2440-2448 (2015)

\bibitem{sermanet12}
Sermanet, P., Chintala, S., LeCun, Y.:
Convolutional neural networks applied to house numbers digit classification.
In: Proceedings of the 21th International Conference on Pattern Recognition (ICPR 2012), IEEE  pp. 3288-3291  (2012)

\bibitem{mnih16}
Mnih, V., Badia, A.P., Mirza, M., Graves, A.,  Harley, T., Lillicrap, T.P., Silver, D., Kavukcuoglu, K.:
Asynchronous methods for deep reinforcement learning.
In: Proceedings of the $33^{rd}$ International Conference on Machine Learning, PLMR 48:1928-1937  (2016)

\bibitem{hwang10}
Wang, H.,  Gao, Y., Chen, X.:
RL-DOT: A reinforcement learning NPC team for playing domination games.
IEEE Transactions on Computational Intelligence and AI in Games 2(1):17-26 (2010)

\bibitem{hausk14}
Hausknecht, M., Lehman, J., Miikkulainen, R., Stone, P.:
A neuro-evolution approach to general Atari game playing.
IEEE Transactions on Computational Intelligence and AI in Games 6(4):355-366 (2014)

\bibitem{diwang15}
Wang, D., Tan, A.H.:
Creating autonomous adaptive agents in a real-time first-person shooter computer game.
IEEE Transactions on Computational Intelligence and AI in Games 7(2):123-128 (2015)

\bibitem{williams92}
Williams, R.J.:
Simple statistical gradient-following algorithms for connectionist reinforcement learning.
Machine Learning, 8(3): 229-256 (1992)

\bibitem{tsitsiklis97}
Tsitsiklis, J.N., Van Roy, B.: 
An analysis of temporal-difference learning with function approximation. 
IEEE Transactions on Automatic Control, 42(5):674–690 (1997)

\bibitem{span15}
Phon-Amnuaisuk, S.:
Evolving and discovering Tetris gameplay strategies.
In: Proceedings of the 19th Annual Conference on Knowledge-Based and Intelligent Information \& Engineering systems (KES 2015), Procedia Computer Science. Volume 60, pp. 458-467 (2015)

\bibitem{silver14}
Silver, D., Lever, G., Heess, N., Degris, T., Wierstra, D., Riedmiller, M.:
Deterministic policy gradient algorithms. In: Proceedings of the International Conference on Machine Learning (ICML). pp. 387-395 (2014)

\bibitem{andry16}
Andrychowicz, M., Denil, M., Colmenarejo, S. G., Hoffman, M. W., Pfau, D., Schaul, T., Shillingford, B., de Freitas, N.: 
Learning to learn by gradient descent by gradient descent. 
Advances in Neural Information Processing Systems. pp. 3981-3989 (2016)

\bibitem{levine16}
Levine, S., Finn, C., Darrell, T., Abbeel, P.:
End-to-end training of deep visuomotor policies.
Journal of Machine learning Research, 17(1):1334-1373 (2016)

\bibitem{schulman15}
Schulman, J., Moritz, P., Levine, S., Jordan, M.I., Abbeel, P.:
High-dimensional continuous control using generalized advantage estimation.
In: Proceedings of the 4th International Conference on Learning Representation. (ICLR 2016) (2016)

\bibitem{panaluke05}
Panait, L. and Luke, S.:
Cooperative multi-agent learning: The state of the art.
Autonomous Agents and Multi-Agent Systems 11(3): 387-434, (2005)

\bibitem{span09}
Phon-Amnuaisuk, S.:
Learning cooperative behaviours in multiagent reinforcement learning.
In: Proceedings of the 16th International Conference on Neural Information Processing (ICONIP 2009), Part I, pp. 570-579 (2009)

\bibitem{zbigniew05}
Zbigniew, S..:
An analysis of island models in evolutionary computation.
In: Proceedings of the International Conference on Genetic and Evolutionary Computation (GECCO 2005), ACM. pp. 386-389  (2005)

\bibitem{fentress76}
Fentress, J.C.:
Simpler Networks and Behaviour. Sinauer Associates Inc. (1976)

\bibitem{getting80}
Getting, P.A., Lennard, P.R., Hume, R.I.:
Central pattern generator mediating swimming in Tritonia. I. Identification and synaptic interactions.
Journal of Neurophysiology, 44(1): 151-164 (1980)

\bibitem{zeiler13}
Zeiler, M.D., Fergus, B.:
Visualizing and Understanding Convolutional Networks.
In: Proceedings of the European Conference on Computer Vision. (ECCV 2014) pp. 818-833 (2013)

\bibitem{yosinsky15}
Yosinski, J., Chune, J., Fuchs, T., Lipsin, H.:
Understanding neural network through deep visualization.
In: Proceedings of the 31st Conference on Machine Learning, Deep Learning Workshop arXiv:1506.06579v1 (2015)

\bibitem{zhou17}
Zhou, B., Bau, D., Oliva, A., Torralba, A.: 
Interpreting deep visual representations via network dissection.
arXiv:1711.05611 (2017)

\bibitem{william88}
Williams, R.J.:
On the use of backpropagation in associative reinforcement learning.
In: Proceedings of the IEEE International Conference on Neural Networks, IEEE Vol I,  pp. 263-270  (1988)

\bibitem{bertsekas96}
Bertsekas, D.P., Tsitsiklis, J.N.: 
Neuro-Dynamic Programming. Athena Scientific (1996)

\end{small}
\end{thebibliography}
\end{document}